\documentclass{article}
\usepackage{spconf,amsmath,amssymb,graphicx}
\usepackage{mathtools}
\usepackage{subfig}
\usepackage{siunitx}
\usepackage{bmpsize}
\usepackage[abs]{overpic}
\usepackage{adjustbox}
\usepackage[labelsep=period]{caption}
\renewcommand{\thetable}{\Roman{table}}

\usepackage{booktabs,array}
\usepackage{multirow}
\usepackage{balance}
\usepackage[usenames,dvipsnames]{color}
\usepackage[colorlinks=true,linkcolor=NavyBlue, citecolor=NavyBlue, urlcolor=NavyBlue]{hyperref}

\makeatletter
\renewcommand\footnotesize{%
	\@setfontsize\footnotesize\@ixpt{11.4}%
	\abovedisplayskip 8\p@ \@plus2\p@ \@minus4\p@
	\abovedisplayshortskip \z@ \@plus\p@
	\belowdisplayshortskip 4\p@ \@plus2\p@ \@minus2\p@
	\def\@listi{\leftmargin\leftmargini
		\topsep 4\p@ \@plus2\p@ \@minus2\p@
		\parsep 2\p@ \@plus\p@ \@minus\p@
		\itemsep \parsep}%
	\belowdisplayskip \abovedisplayskip
}
\makeatother

\title{Adversarial Inpainting of Medical Image Modalities}
\name{Karim Armanious\textsuperscript{1,2}, Youssef Mecky\textsuperscript{1,3}, Sergios Gatidis\textsuperscript{2}, Bin Yang\textsuperscript{1}}
\address{\textsuperscript{1}University~of~Stuttgart,~Institute~of~Signal~Processing~and~System~Theory,~Stuttgart,~Germany\\
	\textsuperscript{2}University~of~T\"ubingen,~Department~of~Radiology,~T\"ubingen,~Germany\\
	\textsuperscript{3}German~University~in~Cairo,~Faculty~of~Information~Engineering~and~Technology,~Cairo,~Egypt
}
\begin{document}
%
\maketitle
\begin{abstract}

Numerous factors could lead to partial deteriorations of medical images. For example, metallic implants will lead to localized perturbations in MRI scans. This will affect further post-processing tasks such as attenuation correction in PET/MRI or radiation therapy planning. In this work, we propose the inpainting of medical images via Generative Adversarial Networks (GANs). The proposed framework incorporates two patch-based discriminator networks with additional style and perceptual losses for the inpainting of missing information in realistically detailed and contextually consistent manner. The proposed framework outperformed other natural image inpainting techniques both qualitatively and quantitatively on two different medical modalities.
	
\end{abstract}
\section{Introduction}
\label{sec:intro}

Medical imaging is a fundamental tool for diagnostic procedures. Nevertheless, causes for missing or incomplete image information in medical scans are manifold including image artifacts (e.g. metal artifacts in CT and MRI), limited field of view, selective reconstruction of acquired data or superposition of foreign bodies in projection methods. It is clear that missing image information cannot be retrieved in a diagnostic sense, meaning that the actual information is lost. However, for image post processing, completing missing information within medical scans is also of interest.

 One example is attenuation correction in PET/MRI, where MR data are used for the estimation of attenuation coefficients. In this case, it is not the detailed local properties of MR data that are required but a more global property. Here, the correction of missing body parts (e.g. due to artifacts or positioning outside the MR field of view \cite{2}) via inpainting can be of high value \cite{1}. In a similar way, inpainting can be advantageous in radiation therapy planning for the correction of MR artifacts before calculation of dose distribution. In general, completion of medical images is of interest whenever automated algorithms for image analysis shall be applied (e.g. for segmentation or classification) that require a complete, artifact-free input. Thus, inpainting can also be part of data curation frameworks in medical imaging.

Current approaches for medical image inpainting rely on texture synthesis \cite{4}, interpolation \cite{5,6}, non-local means \cite{3} and diffusion techniques \cite{7}. These classical approaches face difficulty when inpainting more complex regions such as in medical imaging data.

From another perspective, the inpainting of natural images is a hot topic of research in the computer vision community. This is especially true while utilizing GANs \cite{8}. Context encoders (CE) are one of the most widely used natural image inpainting techniques \cite{9}. They are based on training an encoder-decoder network with an adversarial discriminative network. However, the resultant inpainted regions may not always be consistent with their surrounding regions. The Globally and Locally Consistent Image Completion (GLCIC) builds upon CE by expanding the discriminator network into a multi-scale approach \cite{1011}. This is achieved by fusing the learned discriminative features from a global discriminator network with those from a more local network before a discriminative decision is taken. However, to ensure consistency with the surrounding regions, further post-processing methods and long training durations are recommended. In \cite{12}, instead of post-processing, the consistency of the inpainted regions is improved by separating the two discriminator networks and using a parsing network to enhance the results. Other proposed inpainting techniques include utilizing contextual attention \cite{13}, perceptual loss \cite{14} or super-resolution methods \cite{15} among others \cite{17}. A complete overview of such methods is outside the scope of this work.

In this work, we introduce the topic of medical image inpainting using deep learning techniques. As a baseline, we utilize our recently proposed MedGAN framework for medical image translation tasks \cite{16}. MedGAN is an adversarial framework combining a cascaded U-net generator architecture (CasNet \cite{16,19}) with a new combination of non-adversarial losses. However, we argue that an inpainting task is more challenging than a translation task. This is because not only the inpainted region must be highly realistic, but also it must fit homogeneously into the given context information. Motivated by the recent advances in natural image inpainting \cite{1011,12}, we expand MedGAN with an additional local discriminator network to enhance the inpainting performance.

Our new model, named ip-MedGAN, produces globally consistent and realistic results without the need for further post-processing. We demonstrate the model performance on different medical modalities, MRI and CT. Furthermore, we compare qualitatively and quantitatively with other adversarial inpainting methods.
\section{Methods}
\begin{figure}[!t]
	\centering
	
	\includegraphics[width=0.5\textwidth]{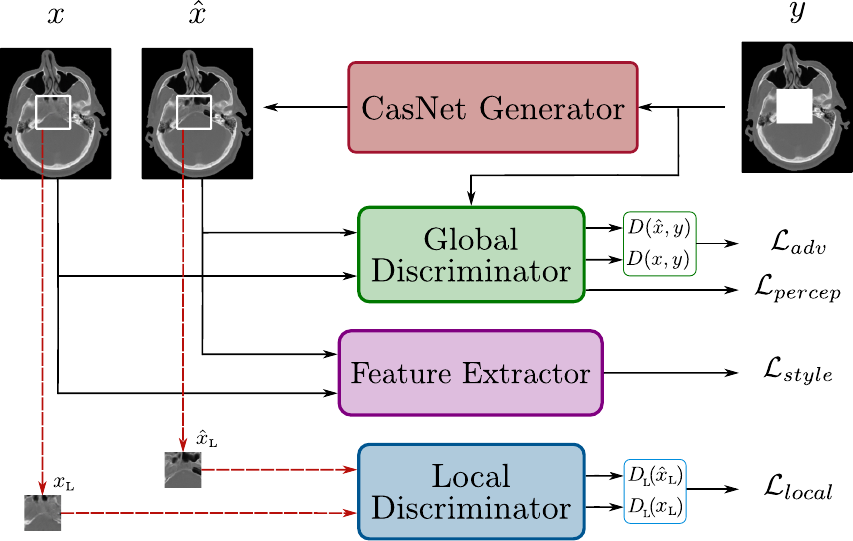}
	
	\caption{An overview of the proposed ip-MedGAN framework utilized for the inpainting of medical image modalities.}
	\label{fig1}
	\vspace{-2mm}
\end{figure}

Our model, ip-MedGAN, is based on a conditional GAN (cGAN) architecture with the inclusion of a patch-based local discriminator network and additional non-adversarial losses. In Fig.\ref{fig1} an outline of the proposed model is presented. \vspace{-2mm}
\subsection{Conditional Generative Adversarial Networks}
A cGAN framework consists of two convolutional networks, the generator $G$ and the global discriminator $D$ \cite{18}. In the proposed framework, $G$ receives as input the context image $y$. It is a 2D medical image of size $256 \times 256$ with a randomly cropped square region of size $64 \times 64$. Thus, the missing portion of the image is $\frac{1}{16}$ of the original image size. The generator utilizes the given context information to inpaint the missing region and to form a synthetically completed image $\hat{x}$. On the other hand side, the discriminator receives as input the target image $x$ with no missing information or the generated image $\hat{x}$. It utilizes a binary cross entropy loss function to classify which of input images is a synthetic output from the generator, $D(\hat{x},y) = 0$, and which belongs to the real target distribution, $D(x,y) = 1$. The networks are trained via a game-theoretical approach where the generator attempts to fool the discriminator into misclassifying $\hat{x}$ as a real image, while the discriminator constantly improves its classification performance to avoid being fooled. The following min-max optimization task represents this training procedure:
\begin{equation}
\min_{G} \max_{D} \mathcal{L}_{\small\textrm{adv}} = \mathbb{E}_{x,y} \left[\textrm{log} D(x,y) \right] + \mathbb{E}_{\hat{x},y} \left[\textrm{log} \left( 1 - D\left(\hat{x},y\right) \right) \right]
\label{e1}
\end{equation}
where $\mathcal{L}_{\small\textrm{adv}}$ is the adversarial loss function.

For the generator network, a CasNet architecture is utilized which cascades multiple U-net networks, with batch normalization and skip-connections, in an end-to-end manner. This is utilized to distribute the generative task over the more extensive network and thus produce more detailed outputs. CasNet has been shown as an effective method of increasing the overall network capacity and stabilizing the training while avoiding depth-related problems such as vanishing gradients and exponential increase in the number of parameters \cite{19}. Further architectural details are presented in \cite{16}.

The discriminator network is identical to the architecture proposed in \cite{20}. It is a patch discriminator which divides the input images into overlapping $70 \times 70$ patches, before classifying each patch as real or fake. For the final classification decision, the score from all patches is averaged out. We consider this discriminator network as a global discriminator because it takes as input the complete output and target images and not just the inpainted and target patches. By focusing on smaller patches which collectively span the entire image, the global discriminator ensures that the inpainted regions fit homogeneously into the given context information. \vspace{-2mm}
\subsection{Patch-Based Local Discriminator}
Inspired by recent natural image inpainting techniques, the proposed model extend MedGAN by including an additional discriminator network titled the local discriminator $D_{\tiny\textrm{L}}$ \cite{1011,12}. In contrast to the global discriminator, $D_{\tiny\textrm{L}}$ receives as input only the inpainted and target regions, $\hat{x}_{\tiny\textrm{L}}$ and $x_{\tiny\textrm{L}}$ respectively. This allows the local discriminator to focus on the details of the inpainted region rather than on the global context information in the complete image. $D_{\tiny\textrm{L}}$ is also a patch-based network which divides the input regions in $34 \times 34$ overlapping patches for classification. It is trained in an adversarial setting along with the generator network analogous to Eq. \ref{e1}:
\begin{equation}
\min_{G} \max_{D_{\tiny\textrm{L}}} \mathcal{L}_{\small\textrm{local}} = \mathbb{E}_{x_{\textrm{L}}} \left[\textrm{log} D_{\tiny\textrm{L}}(x_{\tiny\textrm{L}}) \right] + \mathbb{E}_{\hat{x}_{\tiny\textrm{L}}} \left[\textrm{log} \left( 1 - D_{\tiny\textrm{L}}\left(\hat{x}_{\tiny\textrm{L}}\right) \right) \right]
\end{equation}
\vspace{-7mm}
\subsection{Non-Adversarial Losses}
To improve the inpainted results, additional non-adversarial losses are utilized to train the generator network. The first is the style reconstruction loss which guides the generator to match the style and textures of the target images $x$ onto the generated output $\hat{x}$ \cite{21,22}. This loss is calculated using intermediate features maps extracted from a pre-trained feature extractor network. A VGG-19 network pre-trained on the ImageNet classification task is utilized \cite{23}. The extracted feature maps are used for the calculation of the Gram matrices, $G_n(x)$ and $G_n(\hat{x})$, which represent the correlation between the features in the spacial extend for $x$ and $\hat{x}$, respectively \cite{16}. The style reconstruction loss is calculated as the weighted average of squared Frobenius norm of the Gram matrices: 
 \begin{equation}
\mathcal{L}_{\small\textrm{style}} = \sum_{n = 1}^{N} \lambda_{sn} \frac{1}{4 d_n^2} \lVert{G_n\left(\hat{x}\right) - G_n\left(x\right)}\rVert_F^{2}
\end{equation}
where $d_n$ and $\lambda_{sn} > 0 $ are the spatial depth and the weight, respectively, of the extracted features from the $n^{\textrm{th}}$ layer of the feature extractor network and $N$ is the total number of layers.

The second non-adversarial loss utilized within the framework is the perceptual loss. It focuses on minimizing pixel-wise variations as well as perceptual discrepancies between the output and target images, which results in more globally consistent generated images \cite{24}. To evaluate the perceptual loss, the mean absolute error (MAE) between the image inputs, $x$ and $\hat{x}$, and their intermediate feature maps, extracted from the global discriminator network, is calculated. The perceptual loss is then a weighted average of the MAE: \vspace{-2mm}
 \begin{equation}
\mathcal{L}_{\small\textrm{percep}} = \sum_{n = 0}^{B} \lambda_{pn} \lVert{D_n\left(\hat{x},y\right) - D_n\left(x,y\right)}\rVert_1
\label{7}
\end{equation} 
where $\lambda_{pn} > 0$ and $D_n$ are the weight and the extracted feature maps of the $n^{\textrm{th}}$ layer of the  global discriminator, respectively. $B$ is the total number of layers for the global discriminator and $D_{0}$ represents the raw input images. 
\subsection{The ip-MedGAN framework}
To summarize the proposed framework, ip-MedGAN incorporates a CasNet generator together with a global discriminator, which ensures the homogeneity of the inpainted region with the surrounding context information. The framework additionally utilizes a local discriminator to enhance the details of the inpainted output. The generator also minimizes the perceptual and style reconstruction losses for textural and perceptual refinement. The final loss function is given as:
\begin{equation}
\mathcal{L} = \lambda_1 \mathcal{L}_{\small\textrm{adv}} + \lambda_2 \mathcal{L}_{\small\textrm{local}}  + \lambda_3 \mathcal{L}_{\small\textrm{style}} + \lambda_4 \mathcal{L}_{\small\textrm{percep}}
\end{equation}
where $\lambda_{1}$, $\lambda_{2}$, $\lambda_{3}$ and $\lambda_{4}$ represents the contributions of the different loss functions.
\vspace{1mm}
\section{Datasets and Experiments}
The proposed inpainting framework was evaluated on two different medical modalities, CT and MRI. For CT, a dataset of the brain region from 50 volunteers was collected on a clinical CT scanner (Siemens Biograph mCT). The acquired data was resampled from an original resolution of $0.85 \times 0.85 \times 5\textrm{mm}^{3}$ to $1 \times 1 \times 1\textrm{mm}^{3}$. For the training and validation datasets, two-dimensional slices were extracted and scaled to a matrix-size of $256 \times 256$ pixels, from 40 and 10 volunteers respectively. For MRI, 44 anonymized T2-weighted (FLAIR) data sets of the head region acquired on a 3T scanner were used. The MR data was also resampled to $1 \times 1 \times 1\textrm{mm}^{3}$ and rescaled to 2-D slices of matrix size $256 \times 256$ pixels. Scans from 33 patients were used for training and 11 patients for validation. Randomly placed square patches of size $64 \times 64$ were removed from the datasets to form the model's input context images $y$.

To evaluate the performance of ip-MedGAN, qualitative and quantitative comparisons with other inpainting techniques were carried out. Specifically, we compare against CE and GLCIC \cite{9,1011}. To ensure a faithful representation of the comparison methods, verified open-source implementation were utilized along with the hyperparameters from the original publications \cite{25,26}. We also compare against the MedGAN image translation approach \cite{16}. For the weighting of the different utilized loss functions, $\lambda_{1} = 0.8$, $\lambda_{2} = 0.2$ and $\lambda_{3} = \lambda_{4} = 0.0001$ was used for ip-MedGAN, with the original MedGAN framework utilizing instead $\lambda_{1} = 1.0$ and $\lambda_{2} = 0$. All models were trained for 200 epochs on a single NVIDIA Titan X GPU. Training time was on average 24 hours while inference time is approximately 100 milliseconds. For the quantitative comparisons, several evaluation metrics were used: the Universal Quality Index (UQI) \cite{27}, Structural Similarity Index (SSIM) \cite{28}, Peak Signal to Noise Ratio (PSNR) and the Mean Squared Error (MSE). 
\vspace{1mm}
\section{Results and Discussion}

\begin{table}[!t]
	\caption{Quantitative comparison of inpainting techniques\label{t1}}
	\centering
	\setlength\arrayrulewidth{0.05pt}
	\tiny
	\bgroup
	\def\arraystretch{1.15}
	\resizebox{\columnwidth}{!}{%
		\begin{tabular}{r|cccc}
			\hline\hline
			\multirow{2}{*}{Model} & \multicolumn{4}{c}{(a) CT inpainting}\\ & SSIM & PSNR(dB) & MSE & UQI\\
			\hline
			CE & 0.6235 & 19.07 & 1260.2 & 0.9307\\
			GLCIC & 0.7137 & 22.18 & 1169.1 & 0.9290\\
			MedGAN & 0.8044 & 29.74 & 368.9 & 0.9681
			\\
			ip-MedGAN & \textbf{0.8346} & \textbf{31.45} & \textbf{284.4} & \textbf{0.9737}
			\\
			\hline \hline
			\multirow{2}{*}{Model} & \multicolumn{4}{c}{(b) MRI inpainting}\\ & SSIM & PSNR(dB) & MSE & UQI\\
			\hline
			CE & 0.1383 & 14.29 & 2624.7 & 0.8492\\
			GLCIC & 0.2287 & 15.01 & 2286.6 & 0.8229\\
			MedGAN & 0.3034 & 15.91 & 1809.5 & 0.7830
			\\
			ip-MedGAN & \textbf{0.3818} & \textbf{18.32} & \textbf{1121.2} & \textbf{0.9262}
			\\
			\hline
		\end{tabular}
	}
	\egroup
	\vspace{-5mm}
\end{table}

\begin{figure*}[t]
	\begin{minipage}[t]{1.0\linewidth}
		\centering
		\vspace{7mm}
		\begin{minipage}[t]{0.175\linewidth}
			\centering
			\begin{overpic}[width=0.837\textwidth]%
				{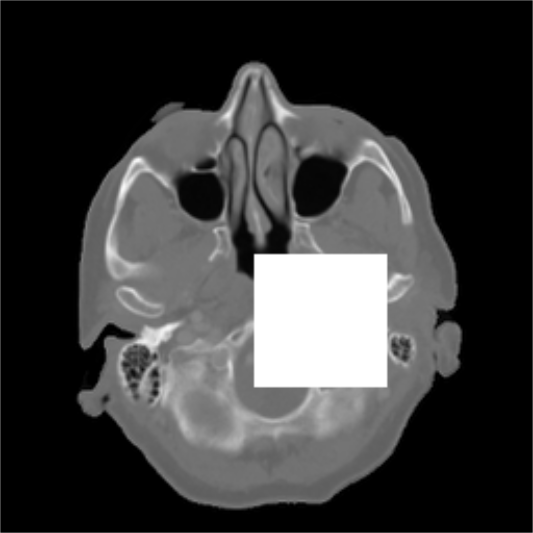}
				\centering
				\put(26,83){Input}
			\end{overpic}	
		\end{minipage}%
		\begin{minipage}[t]{0.645\linewidth}
			\centering
			\begin{overpic}[width=0.227\textwidth]%
				{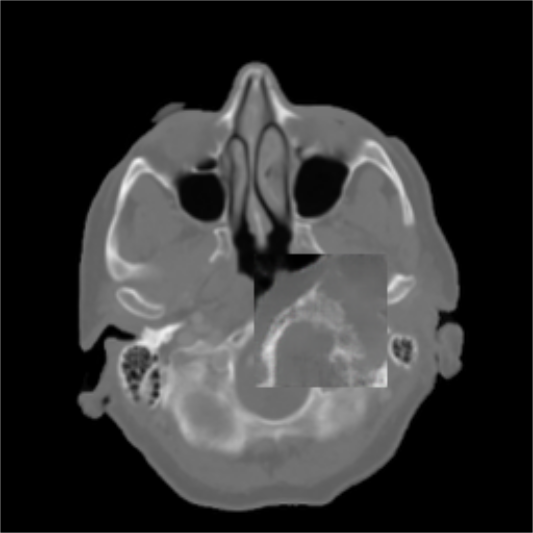}
				\centering
				\put(30,83){CE}
			\end{overpic}
			\begin{overpic}[width=0.227\textwidth]%
				{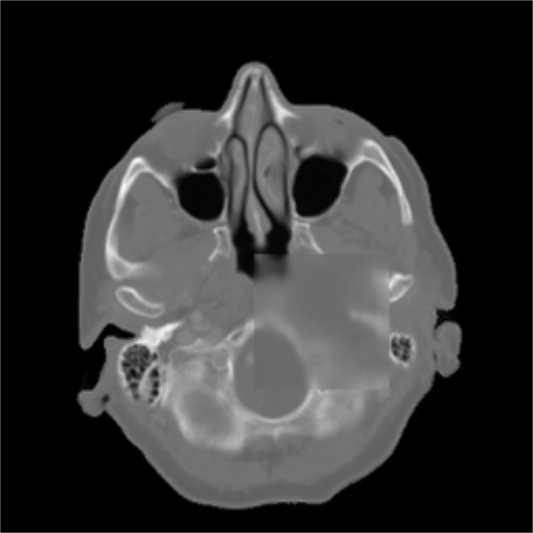}
				\centering
				\put(20,83){GLCIC}
			\end{overpic}
			\begin{overpic}[width=0.227\textwidth]%
				{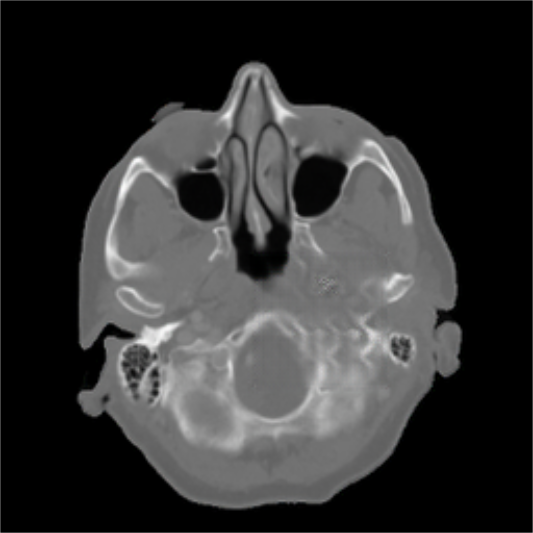}
				\centering
				\put(16,83){MedGAN}
			\end{overpic}
			\begin{overpic}[width=0.227\textwidth]%
				{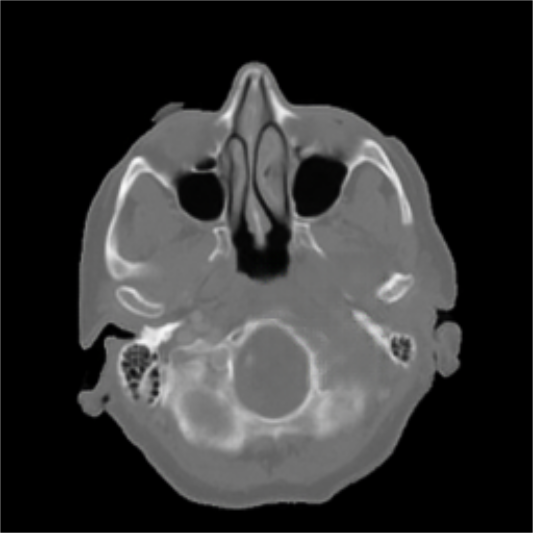}
				\centering
				\put(10,83){ip-MedGAN}
			\end{overpic}
		\end{minipage}
		\begin{minipage}[t]{0.160\linewidth}
			\centering
			\begin{overpic}[width=0.916\textwidth]%
				{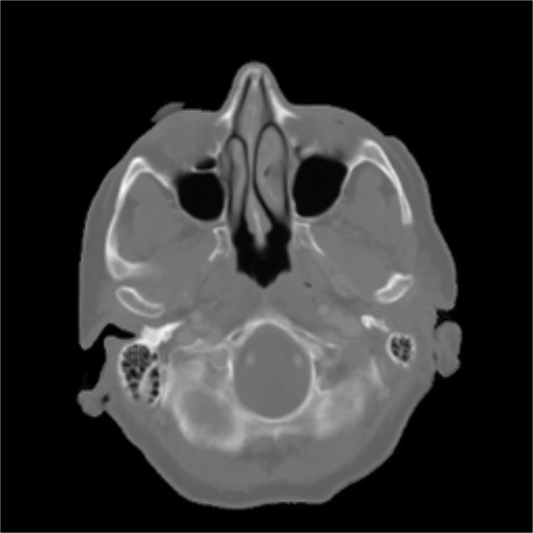}
				\centering
				\put(22,83){Target}
			\end{overpic}		
		\end{minipage}\\
	\vspace{2mm} 
	\begin{minipage}[t]{0.175\linewidth}
		\centering
		\begin{overpic}[width=0.837\textwidth]%
			{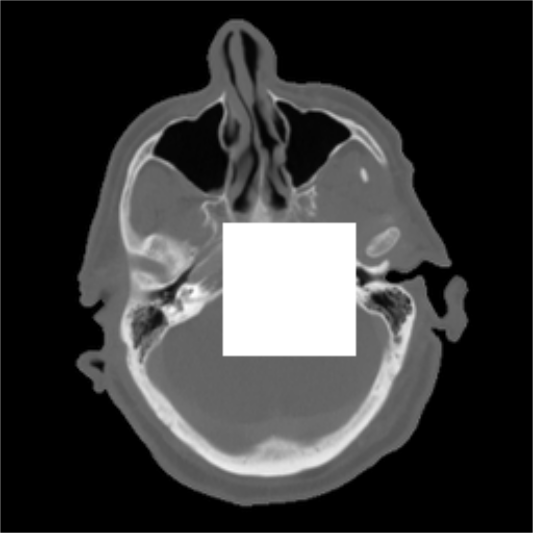}
		\end{overpic}	
	\end{minipage}%
	\begin{minipage}[t]{0.645\linewidth}
		\centering
		\begin{overpic}[width=0.227\textwidth]%
			{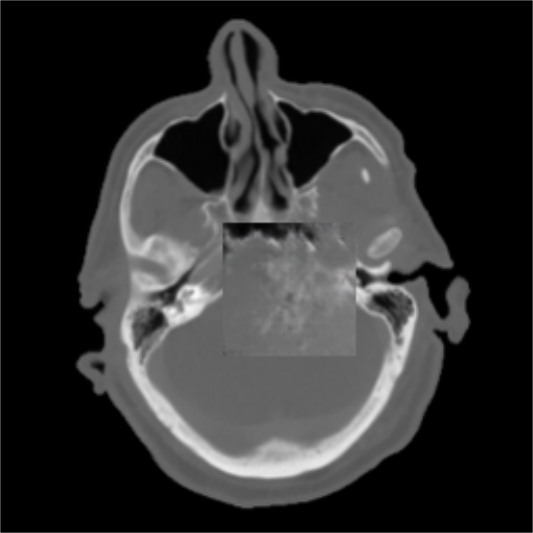}
		\end{overpic}
		\begin{overpic}[width=0.227\textwidth]%
			{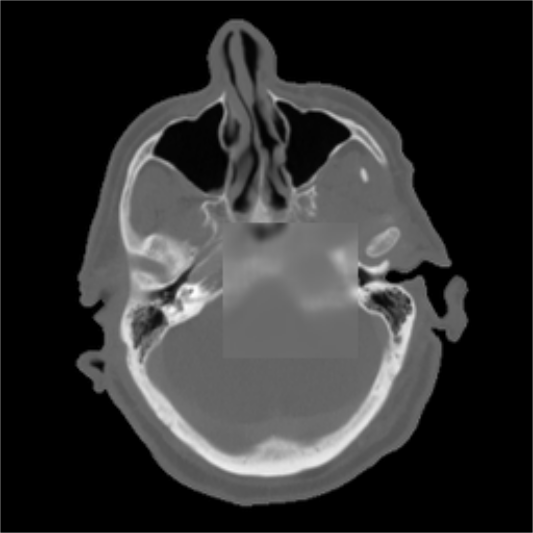}
		\end{overpic}
		\begin{overpic}[width=0.227\textwidth]%
			{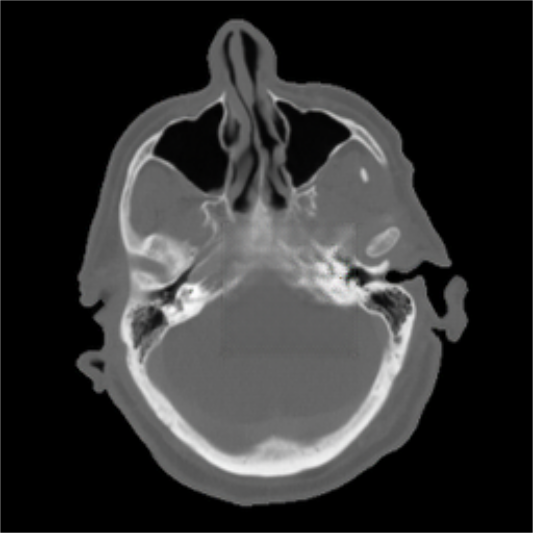}
		\end{overpic}
		\begin{overpic}[width=0.227\textwidth]%
			{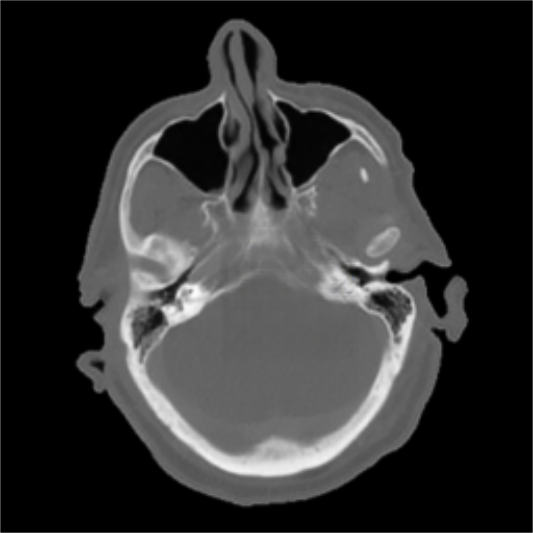}
		\end{overpic}
	\end{minipage}
	\begin{minipage}[t]{0.160\linewidth}
		\centering
		\begin{overpic}[width=0.916\textwidth]%
			{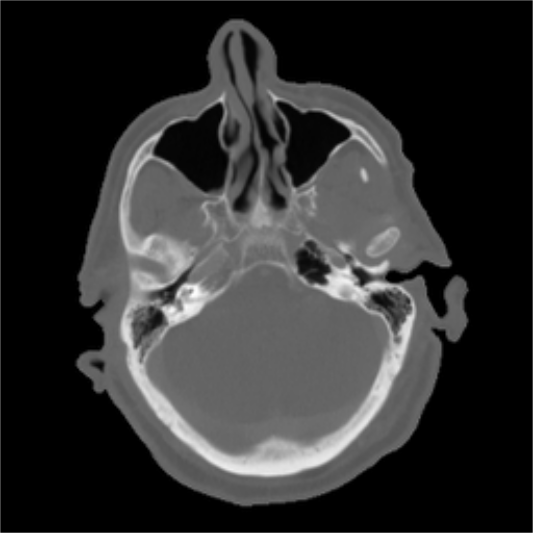}
		\end{overpic}		
	\end{minipage}\\
\vspace{2mm} 
\begin{minipage}[t]{0.175\linewidth}
	\centering
	\begin{overpic}[width=0.837\textwidth]%
		{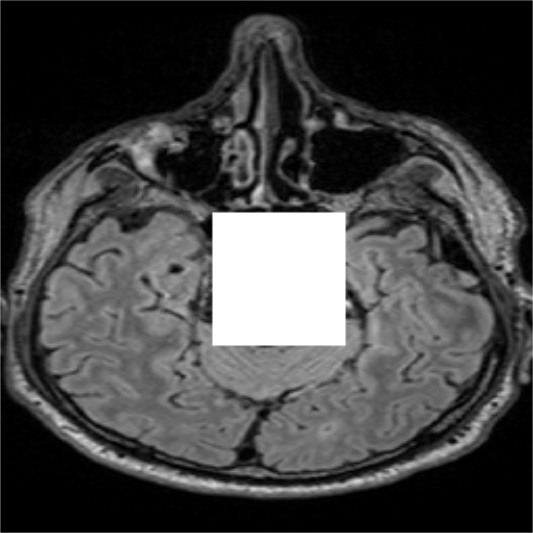}
	\end{overpic}	
\end{minipage}%
\begin{minipage}[t]{0.645\linewidth}
	\centering
	\begin{overpic}[width=0.227\textwidth]%
		{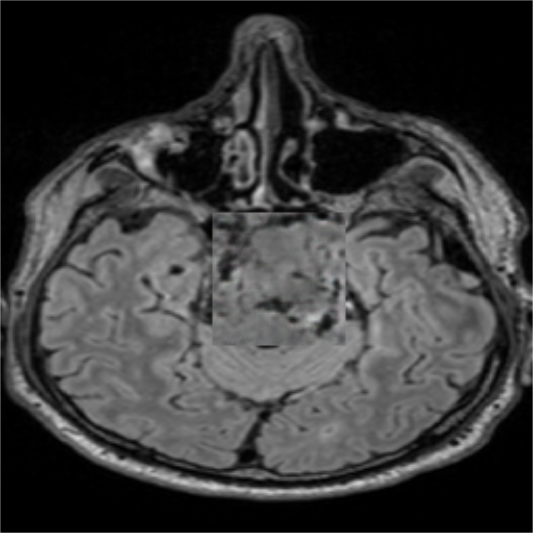}
	\end{overpic}
	\begin{overpic}[width=0.227\textwidth]%
		{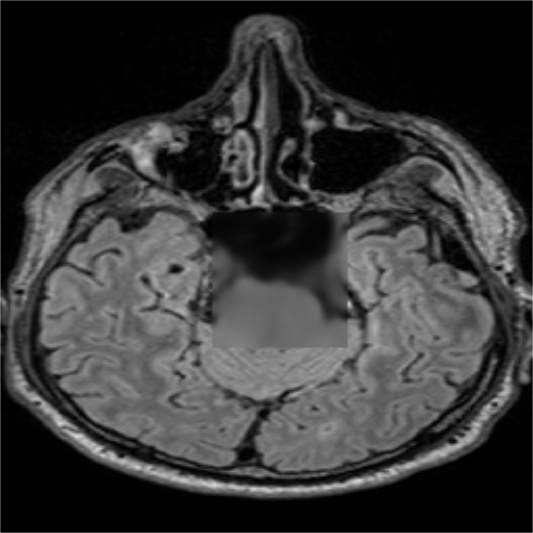}
	\end{overpic}
	\begin{overpic}[width=0.227\textwidth]%
		{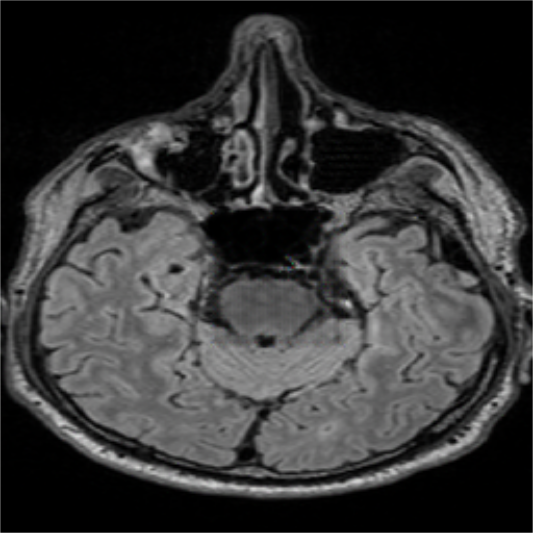}
	\end{overpic}
	\begin{overpic}[width=0.227\textwidth]%
		{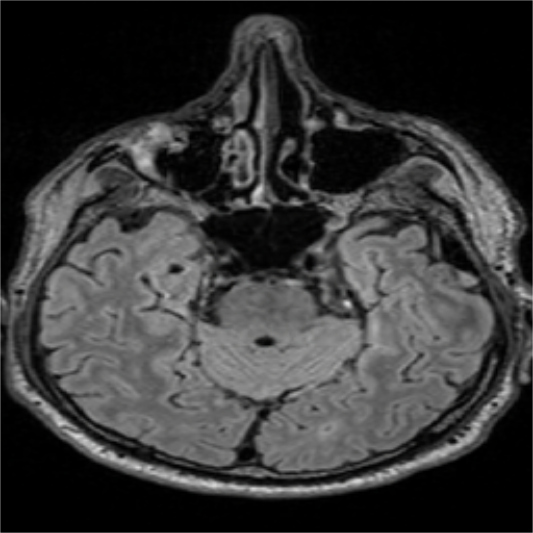}
	\end{overpic}
\end{minipage}
\begin{minipage}[t]{0.160\linewidth}
	\centering
	\begin{overpic}[width=0.916\textwidth]%
		{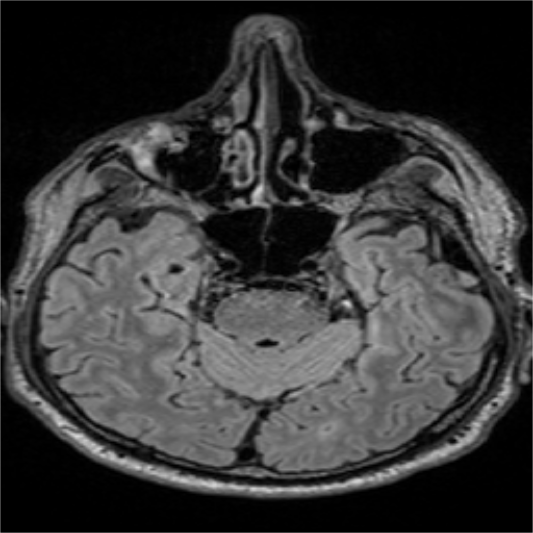}
	\end{overpic}		
\end{minipage}\\
\vspace{2mm} 
\begin{minipage}[t]{0.175\linewidth}
	\centering
	\begin{overpic}[width=0.837\textwidth]%
		{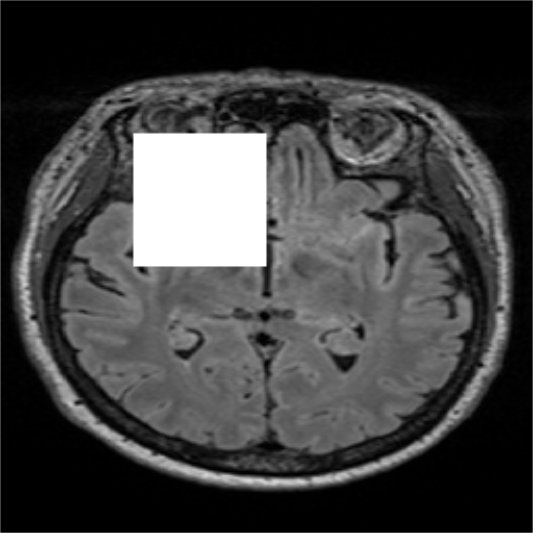}
	\end{overpic}	
\end{minipage}%
\begin{minipage}[t]{0.645\linewidth}
	\centering
	\begin{overpic}[width=0.227\textwidth]%
		{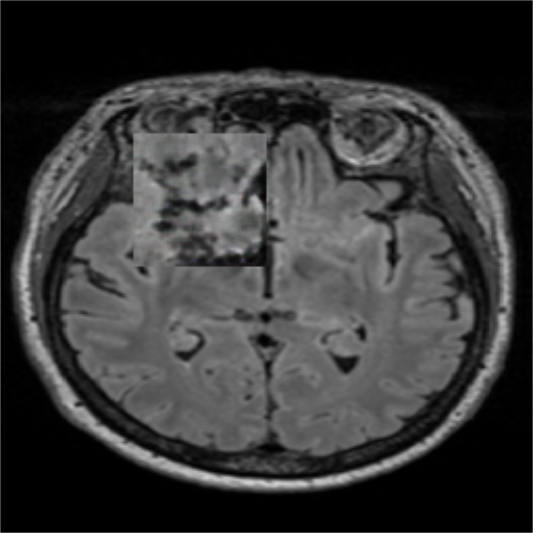}
	\end{overpic}
	\begin{overpic}[width=0.227\textwidth]%
		{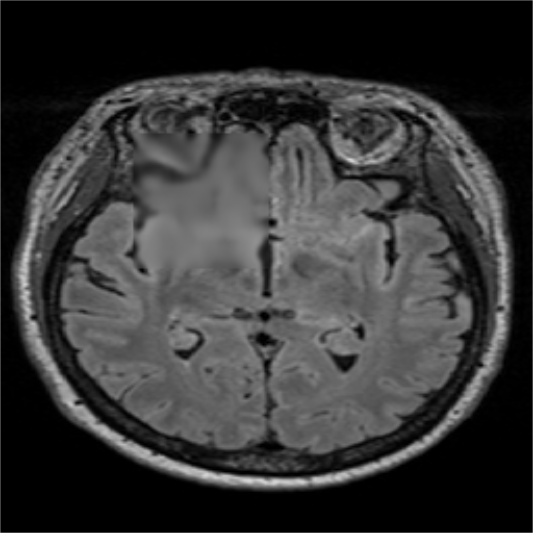}
	\end{overpic}
	\begin{overpic}[width=0.227\textwidth]%
		{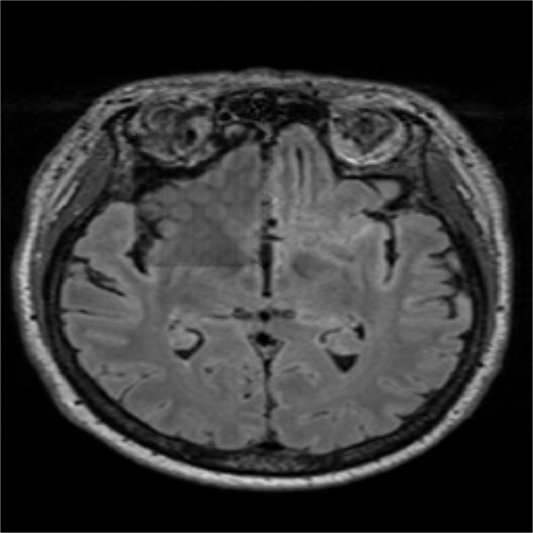}
	\end{overpic}
	\begin{overpic}[width=0.227\textwidth]%
		{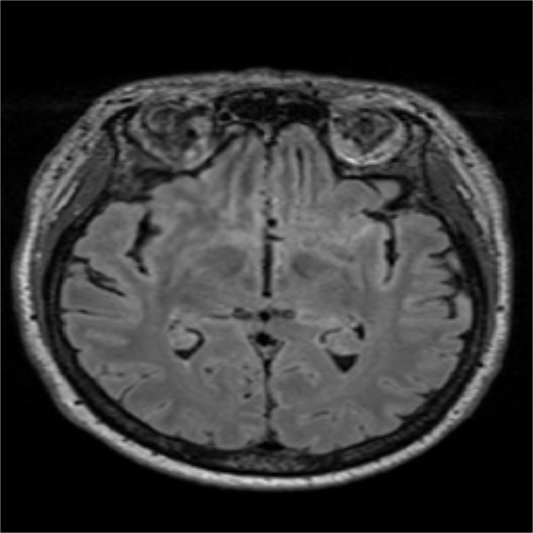}
	\end{overpic}
\end{minipage}
\begin{minipage}[t]{0.160\linewidth}
	\centering
	\begin{overpic}[width=0.916\textwidth]%
		{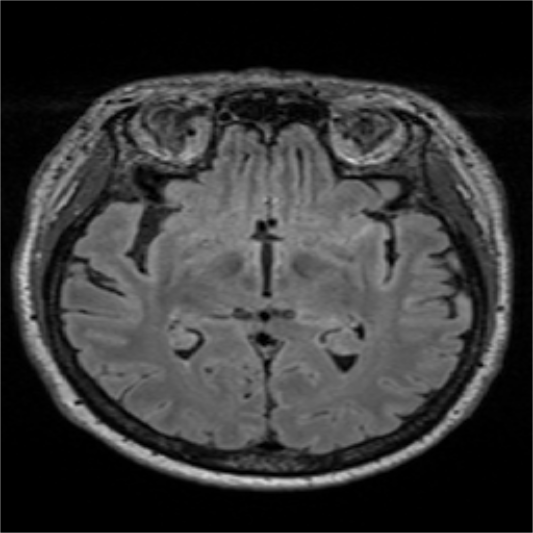}
	\end{overpic}		
\end{minipage}\\
	\end{minipage}
	\caption{Qualitative comparison of the inpainting results between the proposed ip-MedGAN framework and other adversarial inpainting techniques. The first and last two rows represent inpainting of CT and MRI modalities respectively.}
	\label{5}
	\vspace{-4mm}
\end{figure*}

The quantitative and qualitative comparisons of the inpainting performance between the proposed ip-MedGAN framework and other adversarial techniques are presented in Table~\ref{t1} and Fig.~\ref{5} respectively. From a qualitative perspective, CEs produced inpainted regions which did not fit homogeneously into the given context information within the input images. Consequently, this method resulted in the worst quantitative scores in Table~\ref{t1}. The GLCIC framework enhanced the inpainting performance by producing more globally consistent results. However, the inpainted regions were blurry and lacked sharpness. This may be attributed to the relatively short training time, while the original GLCIC paper recommended training for two months on a multi-GPU system with additional post-training image post-processing \cite{1011}. MedGAN produced noticeably enhanced results from the aspect of sharpness and global consistency with the surrounding information. However, the inpainted regions by this method lacked details and contained unrealistic tilting artifacts. By introducing an additional patch-based local discriminator, the proposed ip-MedGAN framework surpasses the limitation of MedGAN by enhancing the textural quality and details of the inpainted regions thus removing any tilting artifacts. This was also reflected quantitatively with the ip-MedGAN framework resulting in the best scores across the chosen metrics.

From another perspective, the proposed ip-MedGAN framework is not without limitation. The training procedure requires the location of the missing regions for the local discriminator. However, this is not necessary for the generator network during inference. This localization may not be readily available in the medical context without the incorporation of an additional segmentation network as a pre-processing step. Moreover, only randomly placed square regions of a fixed size were considered to the input images. However, in the medical context, distortions due to metallic implants in MRI or CT and other similar cases are of arbitrary shapes.
\section{Conclusion}
In this work, we introduce the inpainting of medical images to complete missing or distorted information. This is beneficial for further image post-processing tasks, such as PET/MRI attenuation correction and radiation therapy planning, rather than for diagnostic purposes. To achieve this goal, an adversarial framework is proposed which incorporates two patch-based discriminator networks and additional non-adversarial losses. It ensures that the inpainted results are both detailed and globally consistent in the given context information. The performance of the proposed framework was validated both qualitatively and quantitatively in comparison to other natural image inpainting techniques.

In the future, we plan to expand the proposed framework to include a segmentation network to bypass the need for manual localization of the missing regions during training. Furthermore, we plan to investigate the generalization performance of the proposed model to inpaint arbitrary shapes. Finally, verification of the performance of the inpainted results in further clinical post-processing tasks will be thoroughly investigated in comparison to other traditional approaches \cite{1}.

%


\newpage
\bibliographystyle{IEEEbib}
\balance
{\footnotesize
	\bibliography{refs2}}

\end{document}